\documentclass{article}

\usepackage{arxiv}

\usepackage[utf8]{inputenc} 
\usepackage[T1]{fontenc}    
\usepackage{hyperref}       
\usepackage{url}            
\usepackage{booktabs}       
\usepackage{amsfonts}       
\usepackage{nicefrac}       
\usepackage{microtype}      
\usepackage{graphicx}
\usepackage{natbib}
\usepackage{doi}
\usepackage{lipsum}
\usepackage{graphicx}
\usepackage{subcaption}
\usepackage{amsmath, amssymb}
\usepackage{graphicx}
\usepackage{hyperref}
\usepackage{geometry}
\usepackage[section]{placeins}
\usepackage{float}
\usepackage{algorithm}
\usepackage{algpseudocode}
\usepackage{nameref}
\usepackage{titlesec}
\usepackage{etoolbox}
\usepackage{microtype}
\usepackage[english]{babel} 
\usepackage{amsmath,amssymb,amsfonts}
\usepackage{textcomp}
\usepackage{xcolor}
\usepackage{booktabs}
\usepackage{tabularx}
\usepackage{booktabs}
\usepackage{multirow}
\usepackage{caption}
\usepackage{pdflscape}
\usepackage{rotating}
\usepackage{natbib}
\usepackage{booktabs,tabularx,array,ragged2e,adjustbox}

\title{CDAE: Enhancing Perturbation Robustness in Pretrained Language Models with Contrastive Denoising}

\author{
	1$^{st}$ Sina Heydari \\
	Department of Computer Science and Information Technology \\ 
	Institude for Advanced Studies in Basic Sciences (IASBS) \\
	Zanjan, Iran, 45137-66731\\
	\texttt{sinaheydari@iasbs.ac.ir} \\
	\And
	2$^{nd}$ Amirreza Abbasi\\
	Department of Computer Science and Information Technology \\ 
	Institude for Advanced Studies in Basic Sciences (IASBS) \\
	Zanjan, Iran, 45137-66731\\
	\texttt{a.abbasi@iasbs.ac.ir} \\
	\And
	3$^{rd}$ Mohsen Hooshmand\\
	Department of Computer Science and Information Technology \\ 
	Institude for Advanced Studies in Basic Sciences (IASBS) \\
	Zanjan, Iran, 45137-66731\\
	\texttt{mohsen.hooshmand@iasbs.ac.ir} \\
	\And
	4$^{th}$ Majid Ramezani \\
	Department of Computer Science and Information Technology \\ 
	Institude for Advanced Studies in Basic Sciences (IASBS) \\
	Zanjan, Iran, 45137-66731\\
	\texttt{ramezani@iasbs.ac.ir} \\
}

\begin{document}
	\maketitle
\begin{abstract}
Pre-trained language models have significantly improved sentence representation learning, yet their embeddings remain sensitive to textual perturbations such as synonym substitution, masking, and word dropout. This work proposes a lightweight Contrastive Denoising Autoencoder (CDAE) that refines pre-trained BERT embeddings by jointly optimizing contrastive and reconstruction objectives to learn perturbation-invariant representations. We evaluate the proposed framework under multiple perturbation strategies and strengths, and compare it against the original BERT embeddings and SimCSE. Experimental results show that CDAE consistently preserves higher embedding similarity under perturbation, with the gap widening as the perturbation strength increases, highlighting perturbation-invariant learning as a promising direction for improving the stability of sentence embeddings. 
The source code is publicly available at: https://github.com/ComputationIASBS/CDAE
\end{abstract}


\section{Introduction}
\label{sec:int}
Pre-trained language models embed input text into numerical vectors, to handle downstream tasks~\cite{peters2018}. The resulting vectors represent natural language text as a numerical sequence encoding meaningful semantic information~\cite{devlin2019bert}. Numerous methods have been introduced to obtain effective sentence embeddings, where transformer-based models have become the dominant architecture for learning contextual representations~\cite{tao2024}. A significant advancement was the introduction of the BERT model, which utilizes CLS tokens at the beginning of each sentence. The final hidden representation of CLS token encode information about the entire sentence and are commonly used for downstream tasks~\cite{devlin2019bert}. Despite its pioneering role, BERT performs poorly when it comes to the utilization of sentence embeddings for semantic similarity tasks, presenting challenges for robust sentence representation. To address this limitation, Sentence-BERT (SBERT)~\cite{reimers2019} improved sentence embeddings by adopting a Siamese network architecture and fine-tuning BERT on the SNLI dataset~\cite{bowman2015snli}. In recent years, sentence embeddings have become increasingly important due to the wide range of applications of large language models (LLMs) in information retrieval, recommender systems, and semantic search~\cite{hou2025}. Sentence embeddings are central to these applications because they capture semantic similarity between sentences, enabling more accurate retrieval and comparison of semantic spaces of texts~\cite{tao2024}.

Although sentence embedding models have advanced to produce meaningful semantic representations, a struggle has emerged in the form of adversarial attacks, manipulating machine learning models through carefully crafted inputs. Due to the security implications, substantial research has focused on improving model robustness against such attacks. In many non-conspiratorial practical scenarios, however, the primary challenge is not adversarial attacks but rather natural semantic-preserving perturbations~\cite{zhang2022}. Previous studies have demonstrated that pre-trained models are sensitive to common perturbations such as synonym replacement, word masking, and word dropout, which can substantially alter sentence representations, indicating that there is still potential to enhance model robustness; ideally, robust sentence embeddings should remain stable under small semantic-preserving changes to the input sentence~\cite{morris2020, meng2022, gao2021, hauser2021}.

Most current research emphasizes contrastive learning\cite{gao2021}, denoising~\cite{wang2024}, and robustness-oriented training strategies~\cite{meng2022}. These approaches generally seek to improve semantic alignment, downstream task performance, or robustness against adversarial attacks by learning more discriminative sentence representations~\cite{asl2024}. However, considerably less attention has been devoted to understanding the behavior of pre-trained sentence representations under natural semantic-preserving perturbations and enhancing their stability.

A lack of understanding in this area makes it challenging to identify which components of transformer models are most sensitive to natural linguistic variations. Investigating this problem may yield new insights for improving the robustness of sentence embeddings against semantic-preserving perturbations.

In this study, we investigate the behavior of pre-trained sentence embedding models under semantic-preserving perturbations, revealing a failure to consistently preserve semantic representations when subjected to such perturbations. To address this challenge, we propose CDAE, a Contrastive Denoising Autoencoder that enhances pre-trained models' robustness against perturbations in sentence representation learning. The framework encodes both the original sentence and its perturbed counterpart using a shared frozen pre-trained encoder to obtain their sentence representations. A lightweight refinement network is then trained to preserve semantic consistency and reduce the sensitivity of sentence representations to perturbations.
By operating on top of a frozen pre-trained encoder, the proposed framework introduces only a limited number of trainable parameters.

Building on this framework, the present study:

\begin{itemize}
	\item Differentiates between perturbations and adversarial attacks, emphasizing the improvement of robustness to natural, intrinsic linguistic perturbations.
	\item Conducts a systematic analysis of the stability of pre-trained sentence representations under controlled semantic-preserving perturbations and provides empirical insights into the sensitivity of transformer models.
	\item Proposes a perturbation-invariant sentence representation learning framework that improves the stability of sentence embeddings without modifying the weights of a pre-trained language model.
	\item Conducts a comprehensive evaluation comparing CDAE-generated embeddings with those generated by BERT, SentenceBERT, and SimCSE.
\end{itemize}
The structure of the paper is as follows: Section~\ref{sec:rel} reviews related work in sentence embedding generation; Section~\ref{sec:prop} introduces the proposed method; Section~\ref{sec:res} presents the results; and Section~\ref{sec:conc} concludes the paper.

\section{Related Work}
\label{sec:rel}
This section reviews related work on embedding and representation learning in three categories: sentence embedding learning (\ref{subsec:emblearning}), contrastive and denoising representation learning (\ref{subsec:conderl}), and textual perturbation and robustness (\ref{subsec:pertrob}).

\subsection{Sentence Embedding Learning}
\label{subsec:emblearning}
Pre-trained language models have significantly advanced sentence representation learning by enabling the extraction of contextual semantic information from large-scale corpora~\cite{devlin2019bert}. Early methods, such as SBERT, introduced Siamese and triplet network architectures to generate fixed-length sentence embeddings that preserve semantic similarity and reduce inference costs for retrieval and clustering tasks~\cite{reimers2019}. Subsequent research improved representation quality through contrastive learning objectives and large-scale pretraining strategies, as demonstrated by SimCSE~\cite{gao2021} and the E5 family~\cite{wang2022text} of embedding models. Recent benchmarks, including MTEB~\cite{mteb}, highlight the necessity of evaluating sentence embeddings across diverse downstream tasks rather than relying solely on semantic similarity benchmarks.

\subsection{Contrastive and Denoising Representation Learning}
\label{subsec:conderl}
Contrastive learning has become a dominant paradigm for sentence representation learning by encouraging semantically related sentences to occupy nearby regions of the embedding space while separating unrelated instances. SimCSE~\cite{gao2021} demonstrated that stochastic dropout serves as an effective augmentation strategy for constructing positive training pairs, while supervised variants leverage natural language inference datasets to enhance semantic alignment. Complementary methods have explored denoising objectives, reconstructing corrupted sentences to promote robust latent representations~\cite{wang2021}. More recently, robustness-oriented frameworks have incorporated adversarial perturbations into contrastive objectives to improve embedding robustness against malicious textual attacks~\cite{asl2024}. Although these methods enhance representation quality and downstream performance, they primarily optimize semantic alignment or adversarial robustness rather than directly investigating the stability of pre-trained sentence representations under natural semantic-preserving perturbations. In contrast, CDAE maintains a frozen pre-trained encoder and trains a small refinement network on pooled embeddings, explicitly targeting stability under natural content-word perturbations.

\subsection{Textual Perturbation and Robustness}
\label{subsec:pertrob}
Textual perturbations have been widely examined in the context of adversarial robustness and model evaluation. Existing approaches generate perturbations through lexical substitutions, contextual editing, or adversarial optimization to assess the vulnerability of pre-trained language models and to develop more robust training strategies~\cite{zhang2022}. While these studies offer valuable insights into model robustness against intentionally crafted perturbations, they primarily focus on degrading model performance or defending against adversarial attacks. In contrast, less attention has been given to understanding the effects of naturally occurring semantic-preserving perturbations on pre-trained sentence representations or to explicitly promoting representation stability as a learning objective~\cite{asl2024}. This gap motivates the perturbation-invariant representation learning framework proposed in this work.

\begin{figure*}[htbp]
	\centering
	\begin{minipage}{0.48\linewidth}
		\centering
		\includegraphics[width=\linewidth]{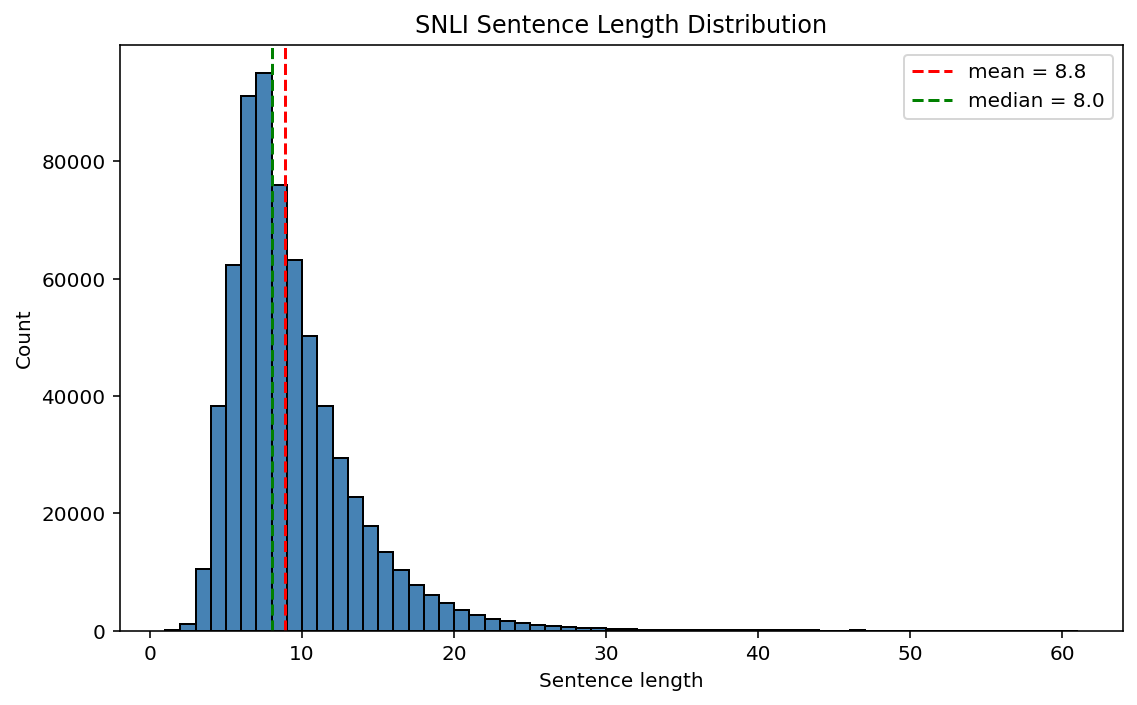}
		\captionof{figure}{Distribution of SNLI sentence lengths after removing the duplicates.}
		\label{fig:sentlength}
	\end{minipage}
	\hfill
	\begin{minipage}{0.48\linewidth}
		\centering
		\includegraphics[width=\linewidth]{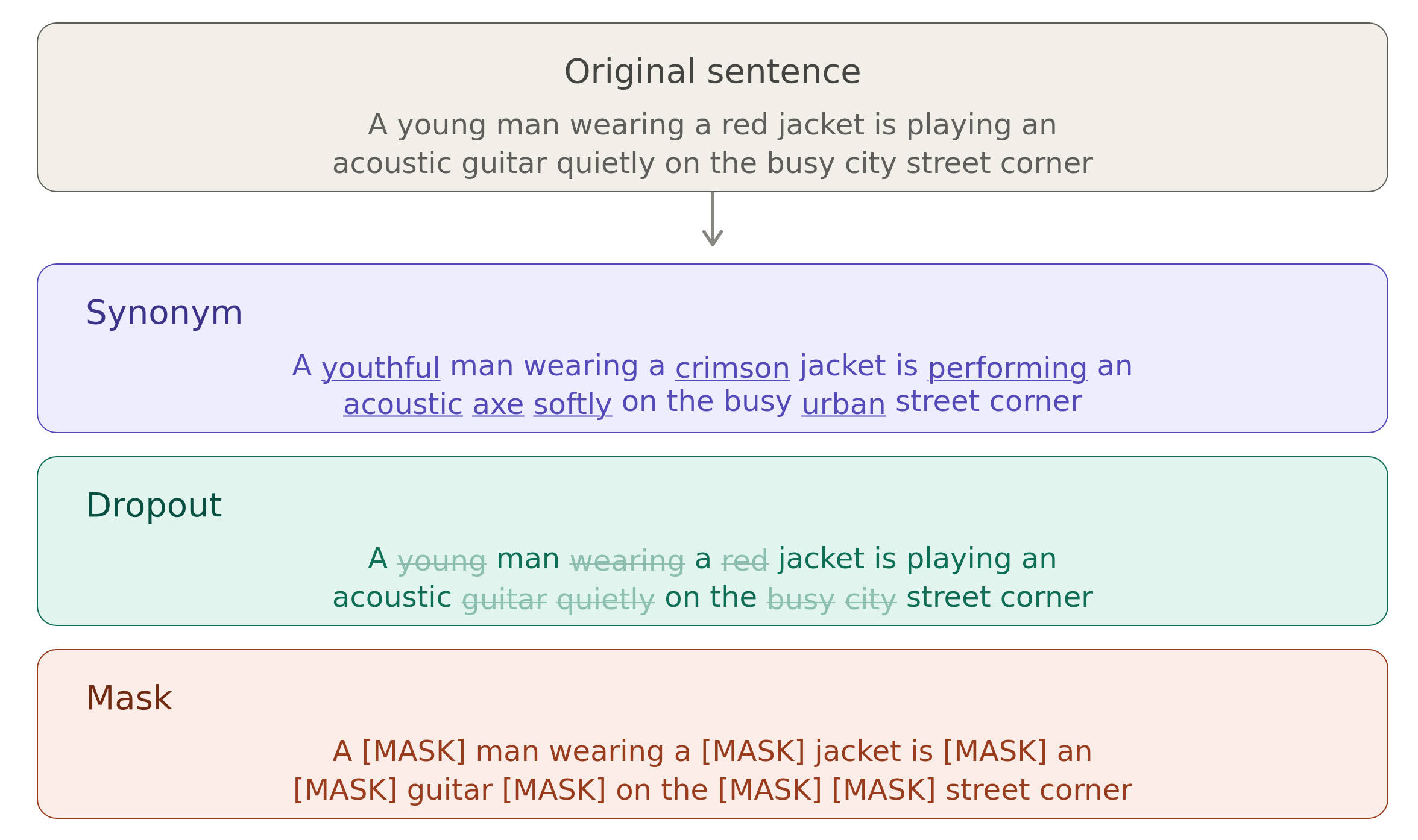}
		\captionof{figure}{Example perturbations applied to an input sentence under each of the three perturbation strategies.}
		\label{fig:pertex}
	\end{minipage}
\end{figure*}

\section{Proposed Method}
\label{sec:prop}
\begin{figure*}[htbp]
	\centering
	\includegraphics[width=\textwidth]{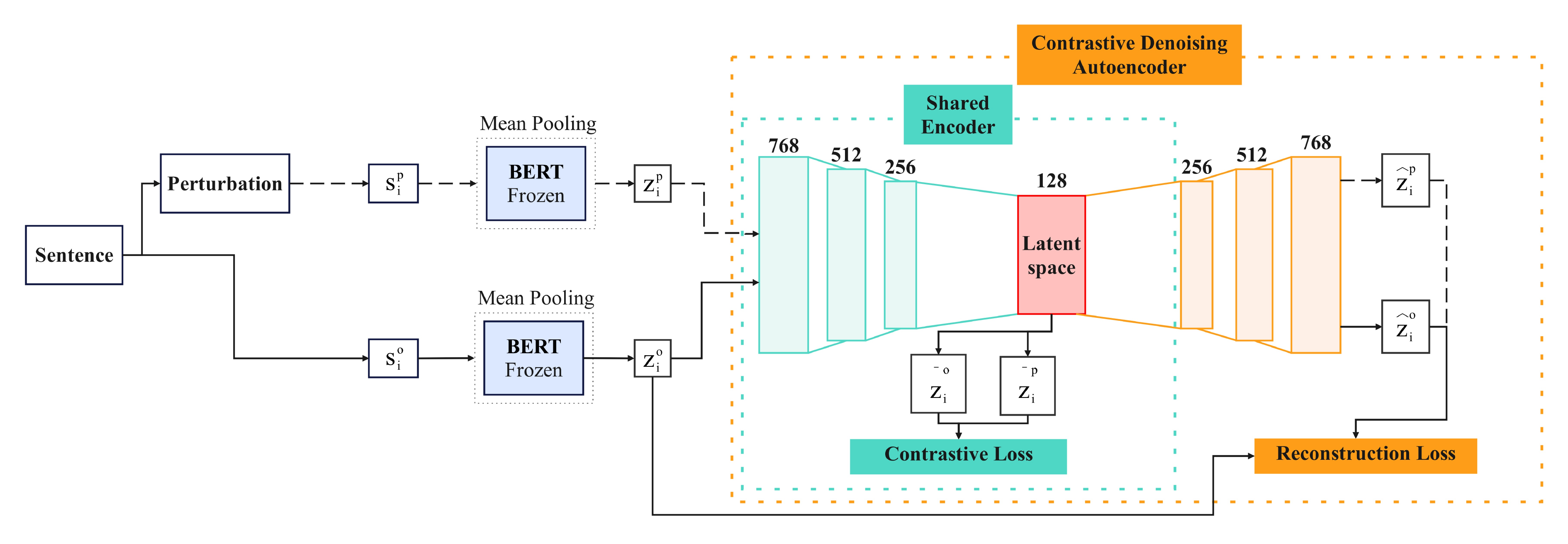}
	\caption{Overall architecture of CDAE, illustrating sentence perturbation, embedding generation, and the flow of original and perturbed representations through the autoencoder. }
	\label{fig:cdae}
\end{figure*}

Pre-trained Language Model (PLM) embeddings often exhibit divergence between perturbed (e.g., synonym-substituted) and original embeddings, which undermines the semantic objectives of large language models (LLMs). To address this challenge, this work introduces the Contrastive Denoising Autoencoder (CDAE), as illustrated in Figure~\ref{fig:cdae}. The following sections first describe the dataset and then detail the CDAE methodology.

\subsection{Dataset}
\label{subsec:data}

The Stanford Natural Language Inference (SNLI) corpus serves as the source of unlabeled natural language sentences for self-supervised representation learning~\cite{bowman2015snli}. While SNLI provides premise-hypothesis pairs annotated with entailment, contradiction, and neutral labels, these annotations are excluded during training. Only premise sentences are used as textual inputs. To prevent redundant samples and bias in contrastive training, only unique premises are retained as training sentences. Following the removal of invalid entries and duplicate (premise, hypothesis, label) rows, 568,484 examples remain across the three splits.

Figure \ref{fig:sentlength} shows the distribution of SNLI sentence lengths after deduplication, with a peak between 7 and 9 tokens per sentence. Most sentences fall between 5 and 15 words (mean 8.8, median 8.0). A long tail extends to 40–60 tokens, representing rare outliers. This concentration of short sentence lengths motivated the selection of a maximum length of 64 during embedding, which covers nearly all examples efficiently and keeps truncation loss negligible.

\subsection{Methodology}

\label{sec:method}

The \emph{CDAE} is designed to learn compact, semantically robust sentence embeddings by combining two complementary training signals on top of a frozen pre-trained sentence encoder. (i) a \textbf{contrastive loss} that aligns a sentence with a lexically perturbed version of itself while separating unrelated sentences within a batch, and (ii) a \textbf{reconstruction loss} that trains a decoder to recover the clean 768-dimensional backbone embedding from the latent code of a perturbed input. These two objectives interact with the autoencoder in distinct ways, as formalized in Section~\ref{subsubsec:gradientflow}. The final sentence representation used for evaluation is the low-dimensional bottleneck vector $\bar{z} \in \mathbb{R}^{128}$.
\subsubsection{Perturbed Sentence Generation}
\label{subsubsec:randomperturbation}

To construct positive pairs for contrastive learning and generate noisy inputs for denoising, CDAE produces, for each sentence $s^i_{o}, i\in {1,\cdots,n}$, a corresponding perturbed variant $s^i_{p}$ using one of the following strategies (Figure~\ref{fig:pertex}): (a) \textbf{Synonym replacement}, where content words are replaced with randomly selected WordNet \cite{miller1992wordnet} synonyms; (b) \textbf{Word dropout}, in which content words are independently removed; and (c) \textbf{Masking}, where content words are replaced with the tokenizer's mask token.

In these strategies, content words—including nouns, verbs, adjectives, and adverbs—are identified using part-of-speech (POS) tagging. Perturbations are controlled by a strength parameter, $\rho \in [0,1]$, which specifies the fraction of eligible content words affected. These are termed content-word perturbations: at low strengths, sentence meaning is largely preserved, while at high strengths, meaning may be altered. Consequently, subsequent evaluation measures representation stability rather than guaranteed semantic equivalence.

\subsubsection{Sentence Embedding}
\label{subsubsec:sentenceembedding}
For each sentence, either $s^o$ or $s^p$, CDAE generates the corresponding embedding using \texttt{bert-base-uncased}~\cite{devlin2019bert} as a frozen backbone, $\text{BERT}: S \rightarrow \mathbb{R}^{768}$, where $S$ denotes the set of all sentences. Given an input sentence $s$, CDAE tokenizes the sentence and obtains contextualized token representations $H \in \mathbb{R}^{L \times 768}$ from the final hidden layer, where $L$ is the sequence length.
The sentence-level embedding is computed by applying mean pooling over the non-padding tokens of the final hidden layer output, as shown below.
\begin{equation}
	\label{eq:frozenbert}
	z =\text{BERT}(s) = \frac{\sum_{j=1}^{L} m_j \cdot h_j}{\sum_{j=1}^{L} m_j},
\end{equation}
where $h_j$ is the contextual representation of token $j$ and $m_j \in \{0, 1\}$ is the attention mask. We write $z^o = \text{BERT}(s^o)$ and $z^p = \text{BERT}(s^p)$ for the clean and perturbed embeddings. 
All backbone parameters remain frozen; thus, $z$ is treated as a fixed input feature rather than a learnable representation.

\subsubsection{CDAE Architecture}
\label{subsubsec:CDAE}
The autoencoder comprises an encoder $Enc(\cdot;\phi)$ and a decoder $Dec(\cdot;\psi)$, both implemented as multilayer perceptrons operating on the fixed 768-dimensional BERT embedding.

\paragraph{Encoder} The encoder transforms the input embedding through hidden layers of size $512 \rightarrow 256$, each followed by batch normalization, ReLU activation, and dropout ($p=0.30$), culminating in a final linear projection to the 128-dimensional latent space $\bar{z}_i = Enc(z_i;\phi) \in \mathbb{R}^{128}$.
Because CDAE employs the InfoNCE loss~\cite{oord2019infonce}, which operates directly on $\bar{z}_i$, the final projection layer does not use an activation function.
\paragraph{Decoder} The decoder mirrors the encoder in reverse, mapping $\bar{z}_i$ through hidden layers of size $256 \rightarrow 512$ to reconstruct the original 768-dimensional space, yielding $\hat{z}_i = Dec(\bar{z}_i;\psi) \in \mathbb{R}^{768}$.
\subsubsection{Loss Functions}
\label{subsubsec:lossfunctions}
\paragraph{Contrastive Loss (InfoNCE)}
For a batch of $B$ sentences, let $\bar{z}^o_i$ and $\bar{z}^p_i$ denote the $\ell_2$-normalized latent codes of the clean and perturbed views of sentence $i$. The perturbed view of sentence $i$ is its positive pair. For the original$\rightarrow$perturbed direction, the perturbed views of the other $B-1$ sentences serve as negatives, and symmetrically the clean views of the other sentences serve as negatives for the perturbed$\rightarrow$original direction. The similarity matrix is shown in \eqref{eq:similaritymatrix}: 
\begin{equation}
	\label{eq:similaritymatrix}
	\text{logits}_{ij} = \frac{(\bar{z}^o_i)^\top \bar{z}^p_j}{\tau},
\end{equation}
where $\tau$ is a temperature hyperparameter. Equation \eqref{eq:contrastiveloss}, shows how we compute a symmetric InfoNCE loss \cite{oord2019infonce} in both directions:
\begin{equation}
	\begin{split}
		\mathcal{L}_{\text{contrast}} =- \frac{1}{2}\Bigg[
		\underbrace{\frac{1}{B}\sum_{i=1}^{B} \log \frac{\exp(\text{logits}_{ii})}{\sum_{j=1}^{B}\exp(\text{logits}_{ij})}}_{\text{original} \rightarrow \text{perturbed}} \\
		+ \underbrace{\frac{1}{B}\sum_{i=1}^{B} \log \frac{\exp(\text{logits}_{ii})}{\sum_{j=1}^{B}\exp(\text{logits}_{ji})}}_{\text{perturbed} \rightarrow \text{original}}
		\Bigg].
	\end{split}
	\label{eq:contrastiveloss}
\end{equation}
No memory bank or momentum encoder is employed; negative samples are drawn exclusively from the current mini-batch.
\paragraph{Reconstruction Loss}
The reconstruction loss in \eqref{eq:reconstructionloss} combines a denoising term and an identity term. Both are measured as mean squared error against the \emph{clean} embedding $z^o$: the denoising term compares the reconstruction of the perturbed input to $z^o$, and the identity term compares the reconstruction of the clean input to $z^o$:
\begin{equation}
	\label{eq:reconstructionloss}
	\mathcal{L}_{\text{recon}} = \underbrace{\frac{1}{d}\left\| \hat{z}^p - z^o \right\|_2^2}_{\text{denoising term}} + \underbrace{\frac{1}{d}\left\| \hat{z}^o - z^o \right\|_2^2}_{\text{identity term}},
\end{equation}
where $d=768$ and the loss is averaged over the batch.
\subsubsection{Combined Objective}
Equation \eqref{eq:totalloss} shows the full training objective, which is a weighted sum of the two losses:
\begin{equation}
	\label{eq:totalloss}
	\mathcal{L}_{\text{total}} = \lambda_{\text{recon}} \, \mathcal{L}_{\text{recon}} + \lambda_{\text{contrast}} \, \mathcal{L}_{\text{contrast}}.
\end{equation}
For all experiments, $\lambda_{\text{recon}} = \lambda_{\text{contrast}} = 1.0$ and $\tau = 0.07$.
\subsubsection{Gradient Flow}
\label{subsubsec:gradientflow}
A central design property of the CDAE is that the two losses reach the shared parameters asymmetrically. Because the backbone is frozen, gradients only propagate through $\phi$ (encoder) and $\psi$ (decoder). Equation \eqref{eq:encodergrad}, shows that the contrastive loss depends solely on the latent codes $\bar{z}^o, \bar{z}^p$ and therefore updates only the encoder:
\begin{equation}
	\label{eq:encodergrad}
	\begin{split}
		\frac{\partial \mathcal{L}_{\text{contrast}}}{\partial \psi} = 0, \\
		\frac{\partial \mathcal{L}_{\text{contrast}}}{\partial \phi} = \frac{\partial \mathcal{L}_{\text{contrast}}}{\partial \bar{z}^o}\frac{\partial \bar{z}^o}{\partial \phi} + \frac{\partial \mathcal{L}_{\text{contrast}}}{\partial \bar{z}^p}\frac{\partial \bar{z}^p}{\partial \phi} \neq 0.
	\end{split}
\end{equation}
The reconstruction loss in \eqref{eq:decodergrad}, in contrast, propagates through the decoder and back through the encoder (since $\hat{z} = Dec(Enc(z;\phi);\psi)$):
\begin{equation}
	\label{eq:decodergrad}
	\begin{split}
		\frac{\partial \mathcal{L}_{\text{recon}}}{\partial \psi} \neq 0,\\
		\frac{\partial \mathcal{L}_{\text{recon}}}{\partial \phi} = \frac{\partial \mathcal{L}_{\text{recon}}}{\partial \hat{z}^o} \frac{\partial \hat{z}^o}{\partial \bar{z}^o}\frac{\partial \bar{z}^o}{\partial \phi} + \frac{\partial \mathcal{L}_{\text{recon}}}{\partial \hat{z}^p} \frac{\partial \hat{z}^p}{\partial \bar{z}^p}\frac{\partial \bar{z}^p}{\partial \phi} \neq 0.
	\end{split}
\end{equation}
Consequently, the encoder receives gradient contributions from both objectives, whereas the decoder is influenced exclusively by the reconstruction objective. This intentional asymmetry allows the latent space to be shaped by an instance-discrimination objective that promotes invariance to perturbation (via InfoNCE), while maintaining sufficient information to reconstruct the original embedding.

\section{Results}
\label{sec:res}

\begin{figure*}[htbp]
	\centering
	\begin{subfigure}{\textwidth}
		\centering
		\includegraphics[width=\textwidth, keepaspectratio]{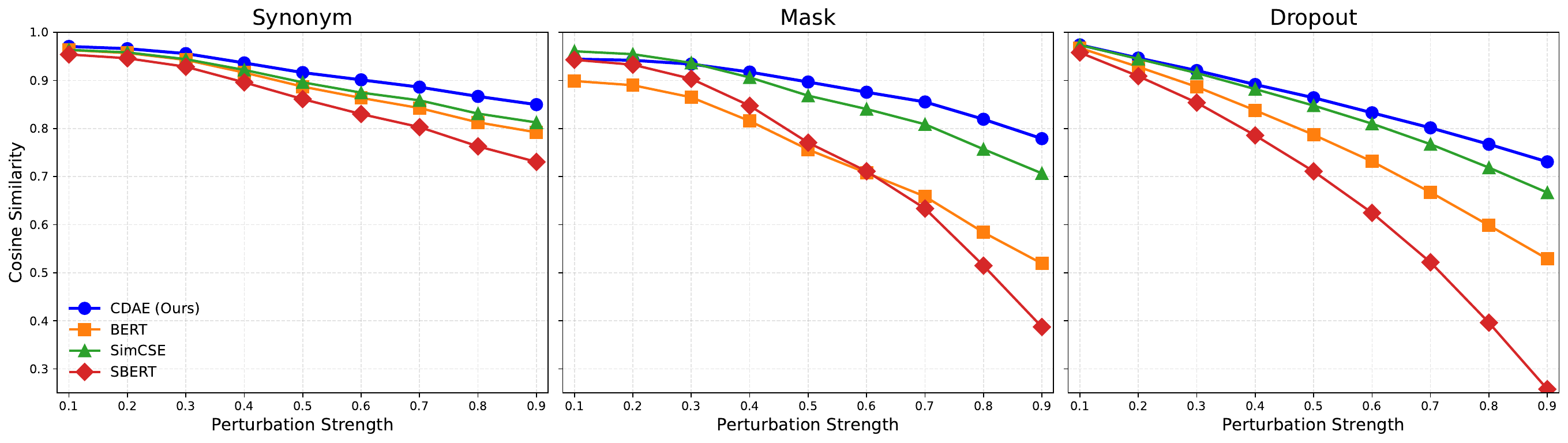}
		\caption{Cosine similarity stability comparative results.}
		\label{fig:pertres}
	\end{subfigure}
	\vspace{0.5em}
	\begin{subfigure}{\textwidth}
		\centering
		\includegraphics[width=\textwidth, height=0.18\textheight]{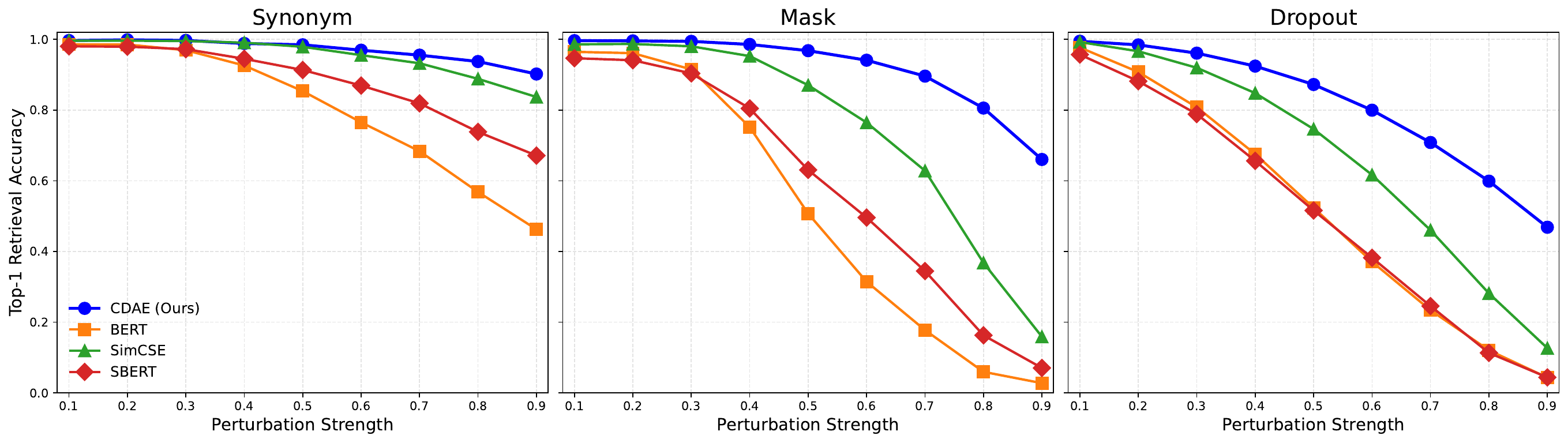}
		\caption{Top-1 retrieval accuracy results comparison.}
		\label{fig:retrieval}
	\end{subfigure}
	\caption{Comparative results of (a) cosine similarity stability and (b) top-1 retrieval accuracy between CDAE and baseline methods under different perturbation strengths.}
	\label{fig:perturbation_comparison}
\end{figure*}

\begin{table*}[htbp]
	\centering
	\small
	\begin{tabular}{c ccc ccc}
		& \multicolumn{3}{c}{Cosine} & \multicolumn{3}{c}{Top-1 Retrieval} \\
		\cmidrule(lr){2-4} \cmidrule(lr){5-7}
		$\rho$ & Drop & Mask & Synonym & Drop & Mask & Synonym \\
		\midrule
		0.1 & .9743 $\pm$ .0003 & .9441 $\pm$ .0007 & .9700 $\pm$ .0002 & .9969 $\pm$ .0010 & .9965 $\pm$ .0009 & .9979 $\pm$ .0005 \\
		0.2 & .9464 $\pm$ .0007 & .9408 $\pm$ .0007 & .9655 $\pm$ .0003 & .9859 $\pm$ .0026 & .9965 $\pm$ .0006 & .9966 $\pm$ .0010 \\
		0.3 & .9182 $\pm$ .0008 & .9319 $\pm$ .0009 & .9558 $\pm$ .0004 & .9583 $\pm$ .0017 & .9957 $\pm$ .0010 & .9983 $\pm$ .0006 \\
		0.4 & .8908 $\pm$ .0012 & .9160 $\pm$ .0014 & .9359 $\pm$ .0007 & .9255 $\pm$ .0018 & .9865 $\pm$ .0028 & .9920 $\pm$ .0009 \\
		0.5 & .8614 $\pm$ .0018 & .8926 $\pm$ .0017 & .9159 $\pm$ .0005 & .8771 $\pm$ .0050 & .9679 $\pm$ .0016 & .9830 $\pm$ .0015 \\
		0.6 & .8290 $\pm$ .0021 & .8734 $\pm$ .0017 & .9012 $\pm$ .0008 & .7967 $\pm$ .0069 & .9468 $\pm$ .0026 & .9747 $\pm$ .0017 \\
		0.7 & .7974 $\pm$ .0026 & .8528 $\pm$ .0021 & .8867 $\pm$ .0007 & .7029 $\pm$ .0052 & .9015 $\pm$ .0027 & .9611 $\pm$ .0011 \\
		0.8 & .7631 $\pm$ .0029 & .8135 $\pm$ .0026 & .8639 $\pm$ .0012 & .5863 $\pm$ .0083 & .7884 $\pm$ .0069 & .9305 $\pm$ .0042 \\
		0.9 & .7287 $\pm$ .0033 & .7748 $\pm$ .0026 & .8484 $\pm$ .0012 & .4705 $\pm$ .0121 & .6593 $\pm$ .0085 & .9079 $\pm$ .0052 \\
		\bottomrule
	\end{tabular}
	\caption{Mean $\pm$ standard deviation of cosine similarity and top-1 retrieval accuracy of CDAE across six random seeds (42, 123, 456, 789, 1024, 1234) for each perturbation method and strength $\rho$.}
	\label{tab:seedresults}
\end{table*}

All experiments were conducted in Python 3.10 using PyTorch \cite{paszke2019pytorch} as the primary deep learning framework and HuggingFace Transformers \cite{wolf2020transformers} for the pre-trained BERT \cite{devlin2019bert} backbone. Experiments were run on a single machine equipped with an NVIDIA GeForce RTX 4060 Laptop GPU (8GB VRAM) and a 13th Gen Intel Core CPU, with all training and inference executed on the GPU (CUDA). The Contrastive Denoising Autoencoder was constructed on a frozen \texttt{bert-base-uncased} backbone and trained for 20 epochs with a batch size of 512, using AdamW \cite{loshchilov2019adamw} as the optimizer.
A learning rate of $1\times10^{-4}$ and weight decay of $1\times10^{-5}$ were used, with a dropout rate of 0.3 applied throughout the encoder and decoder layers to mitigate overfitting. The contrastive temperature was set to $\tau=0.07$, and both loss weights were initialized at 1.0. During training, input sentences were perturbed at a strength of 0.7 (70\% of eligible content words affected) using a randomly selected strategy from synonym replacement, word dropout, or masking to generate the noisy pairs for the denoising and contrastive objectives.

The SNLI dataset \cite{bowman2015snli} was accessed via HuggingFace Datasets \cite{lhoest2021datasets} and processed using pandas \cite{mckinney2010pandas} and NumPy \cite{harris2020numpy}. The training, validation, and test splits contained 548,820, 9,840, and 9,824 examples, respectively, from which unique premise sentences were selected. Sentence perturbation was performed using NLTK \cite{bird2004nltk}, leveraging WordNet \cite{miller1992wordnet} for synonym replacement and its POS tagger for content-word identification. Experiment progress was monitored with tqdm, and robustness results were visualized using Matplotlib.

\subsection{Baselines}
CDAE was compared with embeddings extracted from three pre-trained encoders: BERT (mean pooling), the supervised SimCSE checkpoint \cite{gao2021} (mean pooling), and Sentence-BERT \cite{reimers2019} (mean pooling).

\subsection{Evaluation Protocol}
All test results were obtained from a single evaluation sweep. For each perturbation method $m \in \{\text{synonym}, \text{dropout}, \text{mask}\}$ and strength $\rho \in \{0.1, 0.2, \dots, 0.9\}$, the first $N{=}3000$ unique test premises were selected to prevent data leakage, and one perturbed counterpart of each sentence was generated. The same perturbed sentences were provided to every model, ensuring that differences between models arose solely from their representations. Initially, raw cosine similarity was measured; however, because direct comparisons across embedding spaces can be misleading due to differences in similarity distributions, dimensionality, and representation geometry, top-1 retrieval accuracy was also reported as a complementary, ranking-based metric.

\paragraph{Cosine Similarity} Equation~\ref{eq:similaritytest} shows the mean cosine similarity between clean and perturbed representations:
\begin{equation}
	\label{eq:similaritytest}
	\bar{c}_{m,\rho} = \frac{1}{N}\sum_{i=1}^{N} \cos\!\big(r^o_i,\ r^p_i(m,\rho)\big),
\end{equation}
where $N$ is the number of test sentences, $m \in \{\text{synonym}, \text{dropout}, \text{mask}\}$ is the perturbation method, and $\rho \in \{0.1, 0.2, \dots, 0.9\}$ is the perturbation strength, i.e., the fraction of content words affected. For sentence $s_i$, $r^o_i$ is the representation of the clean sentence and $r^p_i(m,\rho)$ is the representation of its perturbed version $s^p_i$ generated with $(m,\rho)$. The pair $(r^o_i, r^p_i)$ is either the CDAE latent pair $(\bar{z}^o_i, \bar{z}^p_i)$ with $\bar{z}_i = Enc(\text{BERT}(s_i);\phi)$, or the raw embedding pair of a baseline. All models are evaluated on identical perturbed sentences, so differences between models come from the representations only.

Equation~\ref{eq:gap} shows the gap of CDAE over a baseline:
\begin{equation}
	\label{eq:gap}
	\Delta^{b}_{m,\rho} = \bar{c}^{\,\text{CDAE}}_{m,\rho} - \bar{c}^{\,b}_{m,\rho},
\end{equation}
where $b \in \{\text{BERT}, \text{SimCSE}, \text{Sentence-BERT}\}$ is the baseline and $\bar{c}^{\,b}_{m,\rho}$ is computed as in~\eqref{eq:similaritytest} with the raw embeddings of $b$. A positive $\Delta^{b}_{m,\rho}$ means CDAE keeps a sentence closer to its perturbed version than $b$ does at that method and strength.

\paragraph{Top-1 retrieval} Each perturbed sentence $s^p_i$ is used as a query against the $N$ clean sentences $\{s^o_j\}_{j=1}^{N}$, with all vectors taken from the same representation space and ranked by cosine similarity. Top-1 accuracy is
\begin{equation}
	\label{eq:top1}
	\text{Top-1}_{m,\rho} = \frac{1}{N}\sum_{i=1}^{N} \mathbb{1}\Big[\arg\max_{j \in \{1,\dots,N\}}\ \cos\!\big(r^p_i(m,\rho),\ r^o_j\big) = i\Big],
\end{equation}
where $\mathbb{1}[\cdot]$ is the indicator function, equal to 1 if the condition holds and 0 otherwise. The metric measures whether a perturbed sentence stays identifiable among other sentences, which does not depend on how spread out the space is. The chance level is $1/N$.

\subsection{Comparative Results}

\paragraph{Cosine similarity stability}
Figure~\ref{fig:pertres} presents the raw cosine similarity results. CDAE outperforms all three baselines at moderate and high perturbation strengths, with the performance gap increasing as $\rho$ rises. At the lowest strengths, all models perform similarly, with SimCSE slightly surpassing CDAE under masking at $\rho=0.1$ and $0.2$. At $\rho=0.9$, BERT drops to approximately 0.52 under masking and 0.53 under dropout, while SBERT decreases to about 0.39 and 0.26, respectively. SimCSE maintains values around 0.71 and 0.67, whereas CDAE achieves 0.78 and 0.73. Synonym replacement represents the mildest perturbation: CDAE attains 0.85 at $\rho=0.9$, compared to 0.79 for BERT and 0.81 for SimCSE.

\paragraph{Top-1 retrieval accuracy}
Figure~\ref{fig:retrieval} displays top-1 retrieval accuracy. CDAE demonstrates superior performance under masking and dropout across nearly all perturbation strengths, with its advantage increasing as $\rho$ grows. At $\rho=0.9$, CDAE retrieves the correct sentence for approximately 66\% of masked and 47\% of dropout-perturbed inputs, compared to about 16\% and 13\% for the next best model (SimCSE in both cases). Under synonym replacement, all models maintain accuracy close to 1.0 up to $\rho\approx0.3$. BERT and SBERT then decline steadily, reaching approximately 0.46 and 0.67 at $\rho=0.9$. SimCSE closely follows CDAE until $\rho\approx0.5$, after which CDAE maintains a modest lead (approximately 0.90 versus 0.84 at $\rho=0.9$).

Table~\ref{tab:seedresults} reports the stability of CDAE across six random seeds at each perturbation strength. For each seed, CDAE was retrained from scratch, so the reported deviations include variability from weight initialization, mini-batch order, and perturbation sampling. The standard deviation across seeds never exceeds 0.0033 for cosine similarity and 0.0121 for top-1 retrieval, which is far smaller than the differences between methods and between strengths. The largest deviation occurs for dropout at $\rho = 0.9$, where retrieval accuracy is also lowest. Both metrics generally decrease as $\rho$ increases, most steeply under dropout and least under synonym replacement. These results indicate that the reported robustness is not an artifact of a particular seed.

\section{Conclusion}
\label{sec:conc}

This work introduces a lightweight Contrastive Denoising Autoencoder (CDAE) that learns perturbation-robust latent representations through a joint contrastive and denoising objective, thereby improving the stability of pre-trained sentence embeddings under content-word perturbations.

Experimental results indicate that CDAE exhibits greater stability than the original BERT embedding, SimCSE, and Sentence-BERT at moderate and high perturbation strengths. Under masking and dropout, CDAE retrieves the original sentence most reliably among all four models, with the performance margin increasing as perturbation strength rises. The extent to which this gain in stability transfers to downstream semantic tasks was not evaluated in this study and remains an open question.

Future research directions include downstream evaluation, ablation studies of individual loss terms, and training and evaluating CDAE on alternative backbones such as E5. Another avenue is to examine how perturbation sensitivity evolves across different transformer layers and multiple BERT-based models. A layer-wise analysis may yield deeper insights into the establishment of semantic robustness within pre-trained language models and inform the development of perturbation-invariant representation learning methods.

\bibliographystyle{unsrt}
\bibliography{ref} 
\end{document}